\documentclass[10pt,letterpaper]{article}
\usepackage{times}
\usepackage[letterpaper,textwidth=5.5in,textheight=9in,centering]{geometry}
\usepackage[authoryear,round]{natbib}
\setcitestyle{citesep={;},aysep={,},yysep={;}}
\usepackage{enumitem}
\usepackage{float}
\usepackage[utf8]{inputenc}
\usepackage[T1]{fontenc}
\usepackage{url}
\usepackage{booktabs}
\usepackage{array}
\usepackage{amsfonts}
\usepackage{amsmath}
\usepackage{nicefrac}
\usepackage{microtype}
\usepackage{xcolor}
\usepackage{graphicx}
\usepackage[ruled,vlined,noend]{algorithm2e}
\usepackage{placeins}
\usepackage{flafter}
\usepackage[hidelinks]{hyperref}

\hypersetup{
  pdftitle={CuraLoc: Repository-Level Fault Localization via Planned Code Inspection},
  pdfauthor={Sangeeta Warrier, Anush Sankaran, Ryen W. White}
}

\title{CuraLoc: Repository-Level Fault Localization
via Planned Code Inspection}

\author{Sangeeta Warrier \quad Anush Sankaran \quad Ryen W. White\\Microsoft}
\date{}

\begin{document}

\maketitle

\begin{abstract}
A bug report describes what went wrong, but rarely where the code should
change. Existing repository-level localization methods often let language models iteratively decide what code to inspect next, requiring repeated model calls and substantial token usage. We investigate whether these inspection decisions can instead be planned before source retrieval. In this research, we introduce CuraLoc, which first selects candidate files and represents their symbols using compact behavioral and structural descriptions. In a single planning decision, the model jointly selects starting symbols, chooses code relationships to inspect from each, and forms hypotheses about likely edits. Deterministic retrieval collects the requested source for a
final ranking informed by those hypotheses, concentrating model reasoning on choosing and comparing code rather than directing each retrieval step. We demonstrate the efficacy of our proposed approach on the SWE-Bench Lite and Verified datasets. On SWE Bench Verified with GPT-5.2, CuraLoc raises Func@1 from LocAgent's 53.8\% to 59.3\% while using 90.3\% fewer recorded tokens, and also outperforms RepoSearcher (56.2\%) and Agentless (39.6\%) on Func@1. On SWE-bench Lite, CuraLoc achieves a high 83.8\% Func@5 accuracy improving on Agentless by more than 15\% in relative accuracy. Additionally, CuraLoc supports different model backbones, with strong localization accuracy demonstrated using both GPT-5.2 and GPT-6 Astra. These results provide evidence that repository-level code inspection can be planned effectively without iterative model-controlled navigation.
\end{abstract}

\section{Introduction}
\label{sec:introduction}

Automated program repair aims to turn bug reports into
correct code changes. A central challenge is identifying
where those changes belong: a report describes a failure,
but the code that needs to change may lie elsewhere in
the repository. \emph{Fault localization} addresses this
challenge by identifying the files and \emph{symbols}---functions, methods, and classes---that should be
edited~\citep{wong2016survey}.

Code involved in a failure is not necessarily the code
that should change. Consider a library
that rejects a valid input: a public API receives it,
a helper transforms it, and a validator checks it.
If the validator implements the wrong rule, inspecting
its callers can clarify the expected behavior; if the helper
produces the wrong value, inspecting the functions it calls
can clarify the transformation. Each hypothesis about the
edit location suggests a different starting symbol and
a different relationship to inspect.

Existing localization methods handle this choice in two main ways. Fixed-rule methods expand selected locations using a predefined procedure, while interactive agents choose what to inspect next after observing earlier retrievals. The latter can adapt the search to each issue, but requires repeated model calls (\S\ref{sec:related}). 
We ask: \emph{can selecting starting symbols and retrieval
actions from compact code descriptions support accurate
localization without repeated navigation decisions?}

We introduce CuraLoc (\emph{curated localization}). After selecting candidate files from repository structure, CuraLoc gives the primary model compact
symbol descriptions: signatures, structural relationships,
and behavioral summaries from docstrings or a smaller model.
In one decision, the model selects starting symbols,
a retrieval action for each---callers, callees, same-class
methods, or base and overriding implementations---and
hypotheses about likely edits, forming a \emph{RouteMap}.
Deterministic retrieval collects the requested source,
and a final decision ranks edit locations using that
source and the RouteMap's explanations. The inspected
code varies with the issue; the primary model makes
three decisions.

We use SWE Bench Lite for configuration selection and
SWE Bench Verified for primary evaluation~\citep{jimenez2024swebench}.
We compare CuraLoc with Agentless, RepoSearcher, and
LocAgent using GPT-5.2, with three runs per method on
each benchmark. An additional GPT-6 Astra comparison
uses one run per method on SWE Bench Verified.
Func@$k$ requires all reference edit symbols within
the first $k$ distinct predictions.

Our contributions are:
\begin{enumerate}
    \item \textbf{Planned code inspection with model-selected retrieval.}
    We introduce CuraLoc, which uses repository structure
    to select files, compact behavioral and structural
    descriptions to select symbols and related code, and
    retrieved source to rank edit locations. The model
    specifies starting symbols and retrieval actions together;
    deterministic retrieval collects the source for a final
    decision informed by the earlier explanations.
    This design adapts what code is inspected to each issue
    within three primary-model decisions, without repeated
    model-controlled navigation.

    \item \textbf{Accuracy--token gains and tradeoffs.}
    On SWE Bench Verified with GPT-5.2, CuraLoc achieves a Pareto
    improvement over LocAgent: Func@1 rises from 53.8\%
    to 59.3\% while recorded tokens fall by 90.3\%,
    including summary generation. Multiplicity-adjusted
    joint bootstrap bounds support both advantages,
    including on SWE Bench Verified tasks absent from SWE Bench Lite.
    On SWE Bench Lite, comparisons with Agentless and RepoSearcher
    show higher observed symbol accuracy at greater
    token consumption. On SWE Bench Verified, CuraLoc also achieves higher Func@1 than RepoSearcher and Agentless, using $1.74\times$ RepoSearcher's recorded tokens.

\end{enumerate}
Component and failure analyses explain how code selection,
source inspection, and expansion contribute to localization,
and identify where complete edit-set recovery still fails.

\section{Related Work}
\label{sec:related}

\paragraph{Fault localization and interactive search.}
Classical fault localization uses execution coverage and test
outcomes~\citep{wong2016survey,jones2002visualization,abreu2007accuracy},
while information-retrieval methods match bug reports to source
files~\citep{zhou2012buglocator}. Recent methods investigate
repositories through interactive tools in SWE-agent
\citep{yang2024sweagent}, structured search in AutoCodeRover
\citep{zhang2024autocoderover}, graph-guided navigation in
LocAgent~\citep{chen2025locagent}, and priority-based exploration
in OrcaLoca~\citep{yu2025orcaloca}. RepoSearcher alternates
reasoning and retrieval; ToolTrain trains models for this
agent through supervised fine-tuning and tool-interactive
reinforcement learning~\citep{ma2025toolintegratedreinforcementlearningrepo}.
Our comparisons use RepoSearcher with the common primary
model rather than ToolTrain-trained checkpoints.
SHERLOC structures iterative exploration around diagnostic
hypotheses and supplies locations and findings to downstream
repair agents~\citep{sherloc}. These methods let model
decisions respond to observations during the investigation.

\paragraph{Hierarchical localization and graph context.}
Agentless uses hierarchical localization before patch
generation and validation~\citep{xia2024agentless}.
RGFL adds issue-conditioned explanations for candidate
files and code elements, followed by explanation-based
ranking~\citep{sepidband2026rgfl}.
RepoGraph supplies repository-level code relationships
to both hierarchical and interactive methods~\citep{repograph}.
Its Agentless integration retrieves reference-based context
for model-selected locations before source-based localization.
CuraLoc instead asks the model to select both starting symbols
and a retrieval action for each, choosing which relationships
to inspect rather than applying the same retrieval procedure
to every selected location.

\paragraph{Planning and model context.}
ReAct interleaves reasoning, actions, and
observations~\citep{yao2023react}.
ReWOO separates planning, tool execution, and final
synthesis~\citep{rewoo}; LLMCompiler schedules tool calls
according to dependencies, enabling parallel
execution~\citep{llmcompiler}.
CuraLoc studies this separation for localization:
compact symbol descriptions support retrieval choices,
while selected source code supports final ranking.

\section{Proposed Approach}
\label{sec:method}

Given a bug report $d$ and the repository snapshot
$\mathcal{R}$ before the fix, CuraLoc returns up to ten
ranked edit symbols, each identified by its file and
qualified name. An abstract syntax tree (AST)-derived
graph $G$ records symbols, source locations, and code
relationships at that snapshot.

CuraLoc narrows the search in three stages, using more detailed code information at each stage. Its three primary-model
decisions select files, choose symbols and retrieval actions,
and rank edit locations
(Figure~\ref{fig:CuraLoc-pipeline};
Algorithm~\ref{alg:CuraLoc}).
Implementation details appear in
Appendix~\ref{app:method-details}.

\begin{figure}[!htbp]
\centering
\includegraphics[width=.85\linewidth]{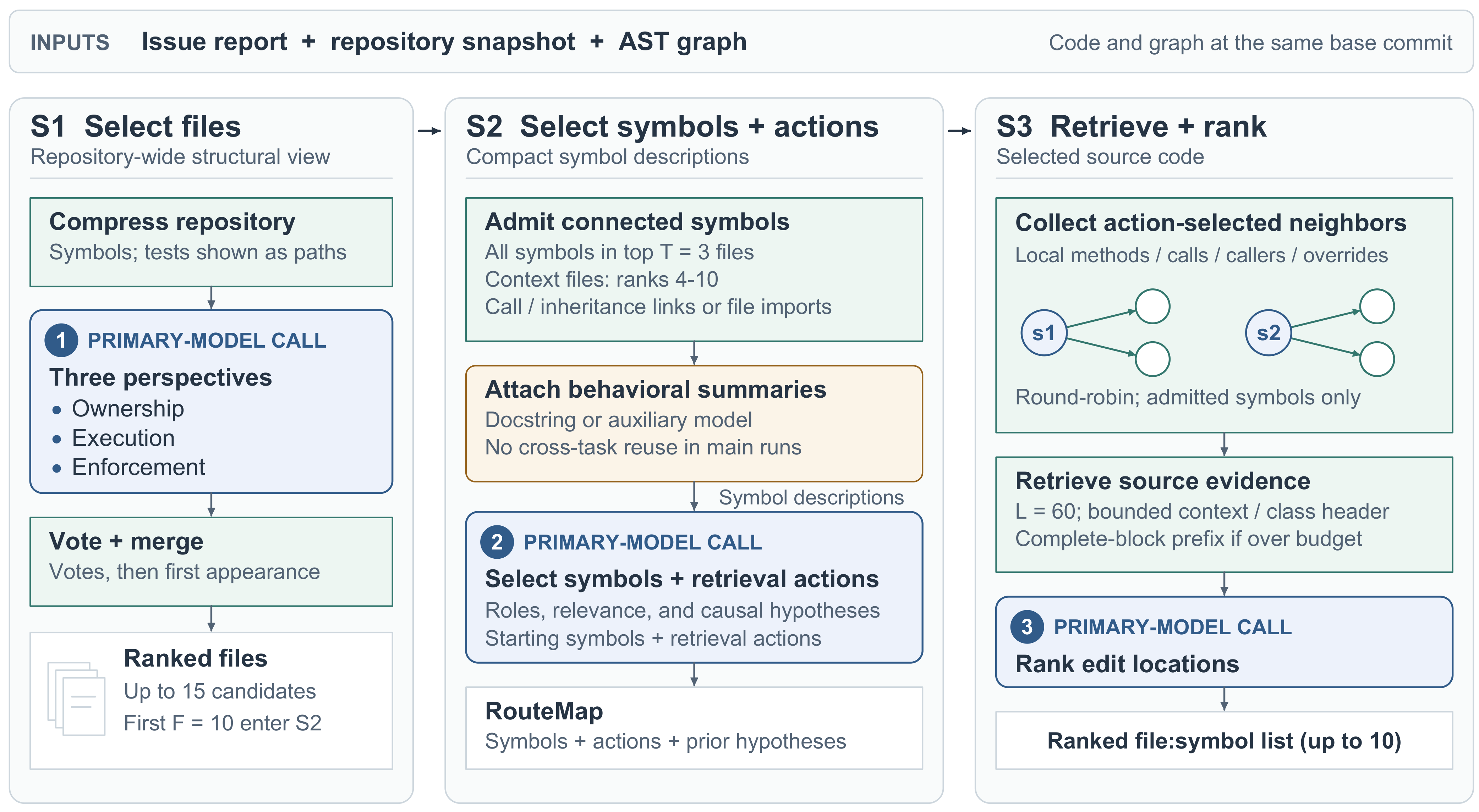}
\caption{CuraLoc uses three primary-model decisions.
Symbol selection specifies both starting locations and
retrieval actions; deterministic retrieval supplies
source code for final ranking.}
\label{fig:CuraLoc-pipeline}
\end{figure}

\begin{algorithm}[t]
\SetAlgoLined
\KwIn{Bug report $d$, repository snapshot $\mathcal{R}$,
AST graph $G$; candidate-file limit $F$, fully included
file count $T\leq F$, source-window limit $L$,
input-token budget $B$}
\KwOut{Ranked edit symbols $\hat{S}$}

\tcc{S1: file selection --- decision 1}
Query the model for ownership, execution, and enforcement
file lists\;
Merge by perspective votes and first appearance;
supplement short lists with bounded structural neighbors\;
\BlankLine

\tcc{Prepare compact symbol descriptions}
Select candidate symbols under the $F,T$ policy;
attach structural information and issue-independent
behavioral summaries\;
\BlankLine

\tcc{S2: select symbols and actions --- decision 2}
Query the model for symbol assessments, ordered starting
symbols, retrieval actions, and hypotheses\;
Validate the RouteMap against the candidate symbols\;
\BlankLine

\tcc{S3: source retrieval and ranking}
Collect starting symbols, then action-selected one-hop
neighbors in round-robin order; restrict to candidate
symbols and deduplicate\;
Retrieve bounded source windows and attach the
RouteMap annotations\;
If input exceeds $B$, retain the longest complete prefix
of source-evidence blocks that fits, without backfilling\;
Query the model to rank up to ten symbols represented
in the prompt \tcp*{decision 3}
Deduplicate predictions, attach available AST metadata,
and return $\hat{S}$\;

\caption{CuraLoc chooses symbols and retrieval actions before
source-based ranking.}
\label{alg:CuraLoc}
\end{algorithm}

\subsection{File Selection (S1)}
\label{sec:stage1}

One model call receives the bug report and compressed
repository structure. It proposes files from three
perspectives: \emph{ownership}, which component defines
the behavior; \emph{execution}, which code participates
in the failing path; and \emph{enforcement}, where
a correction belongs. These perspectives retain different
possible edit locations before detailed inspection.

Files are ranked by the number of perspectives proposing
them, with first appearance breaking ties. The merged list
contains at most 15 files; short lists may be supplemented
with bounded structural neighbors. Rationales and annotations
distinguishing feature-specific consumers from shared
mechanisms accompany the files into S2.

\subsection{Symbol and Action Selection (S2)}
\label{sec:stage2}

\paragraph{Candidate symbols and descriptions.}
CuraLoc considers the first $F$ candidate files.
The candidate set $C$ includes every symbol from the
top $T$ files. For each of the remaining \(F-T\) files, it includes only symbols connected to a symbol in the top \(T\) files through calls, callers, inheritance, or overrides. If one of these files imports, or is
imported by, a top-$T$ file, all its symbols are included.
Files below rank $F$ contribute no symbols.

Each symbol is described by its identity, signature,
structural role, related symbols, and a behavioral summary.
Summaries come from docstrings or a smaller model,
independently of the issue. They can be cached, but main
experiments disable reuse across tasks and runs.
The primary S2 call receives no source bodies.

\paragraph{Choosing what to retrieve.}
One model call produces a \emph{RouteMap}
$\mathcal{M}=(Q,H,P)$.
Here, $Q$ contains symbol scores, issue-specific roles,
and rationales; $H$ contains issue-level hypotheses. Formally, $P=((s_i,a_i,h_i))_{i=1}^{k}$, where each tuple contains a starting symbol $s_i\in C$, a retrieval action
$a_i\in\mathcal{A}$, and hypotheses $h_i$ about the likely edit.

The actions are \texttt{follow\_called\_by} (callers),
\texttt{follow\_calls} (callees),
\texttt{expand\_local} (same-class methods), and
\texttt{check\_override} (base and overriding implementations).
Let $N_a(s)$ denote the symbols returned by action $a$
from $s$ using $G$. Before ordering and input-budget
truncation, the selected evidence symbols are
\[
E_{\mathcal{M}}
=
\bigcup_{i=1}^{k}
\left(
\{s_i\}\cup\bigl(N_{a_i}(s_i)\cap C\bigr)
\right).
\]
S2 thus chooses retrieval actions within the candidate
set established after S1.

The S1 perspectives and S2 roles describe where a fix
may belong---defining behavior, enforcing an invariant,
or accompanying another edit---rather than only which
code is relevant. Their explanations accompany the
source into final ranking. Section~\ref{sec:ablations}
compares this responsibility-oriented formulation with
relevance-oriented guidance;
Appendix~\ref{app:role-taxonomy} defines the roles.
Invalid identifiers are discarded. Malformed plans stop
the current attempt for recovery; finalized failures
count as misses.

\subsection{Source Retrieval and Ranking (S3)}
\label{sec:stage3}

S3 collects starting symbols first, then action-selected
neighbors in round-robin order, without recursive expansion.
Each symbol contributes a primary source window of at most
$L$ lines, with bounded additional class context for methods.
The final prompt combines the issue, source, and RouteMap
annotations, distinguishing S2-assessed symbols from
additional related code. One model call ranks up to ten
edit locations represented in the prompt.
Ordering details, budget truncation, and output handling
appear in Algorithm~\ref{alg:CuraLoc} and
Appendix~\ref{app:method-details}.

\section{Experimental Results}
\label{sec:results}

\subsection{Experimental Setup}
\label{sec:reporting-protocol}

\paragraph{Tasks and methods.}
We select configurations on SWE Bench Lite and evaluate
primarily on SWE Bench Verified~\citep{jimenez2024swebench}.
The cohorts contain 274 and 454 tasks, respectively,
with pre-existing function- or class-level edit targets;
370 evaluation tasks are absent from SWE Bench Lite.
Using GPT-5.2, we compare CuraLoc with
LocAgent~\citep{chen2025locagent},
RepoSearcher~\citep{ma2025toolintegratedreinforcementlearningrepo},
and Agentless~\citep{xia2024agentless} on both benchmarks,
with three runs per method. Methods retain their sampling
and aggregation procedures
(Appendix~\ref{app:eval-details}).
The GPT-6 Astra comparison uses one SWE Bench Verified
run per method for all four methods.

\paragraph{Metrics and computation.}
Func@$k$ requires all reference edit symbols within
the first $k$ distinct normalized predictions.
File@$k$ requires all reference edit files within
the first $k$ distinct files derived from that ranking.
MRR averages first-hit reciprocal rank; finalized
execution failures count as misses.
Only the 335 single-symbol tasks in SWE Bench Verified
can satisfy Func@1, but all 454 remain in its denominator.
Recorded tokens include input and output from successful
responses in retained executions across all contributing
stages and samples. CuraLoc includes summary generation
with cross-task and cross-run summary reuse disabled
(Appendix~\ref{app:timing-details}).

\paragraph{Uncertainty.}
Main comparisons report three-run means and sample
standard deviations. Paired task-bootstrap intervals
use 50,000 resamples, keeping each task's three
executions together; whole-repository resampling
assesses sensitivity to repository composition.
Joint accuracy--token bounds adjust across nine
GPT-5.2 comparisons: three baselines on each benchmark
and on SWE Bench Verified tasks absent from SWE Bench Lite
(Appendix~\ref{app:statistics}).

\subsection{Localization Accuracy and Token Consumption}
\label{sec:main-results}
\label{sec:cost}

\begin{table}[!htbp]
\centering
\caption{GPT-5.2 localization on SWE Bench Verified ($n=454$).
Entries are three-run means with sample standard deviations.
Accuracy is in percent; tokens are in thousands per task.
Bold marks the best reported mean.
Agentless token totals are unavailable.
RepoSearcher returns at most five symbols, so Func@10 equals Func@5.}
\label{tab:method-comparison}
\small
\setlength{\tabcolsep}{3pt}
\begin{tabular}{lrrrrrrr}
\toprule
Method & Func@1 & Func@5 & Func@10 & MRR
       & File@1 & File@10 & Tokens (K) \\
\midrule
CuraLoc
    & $\mathbf{59.3}_{\pm0.9}$
    & $\mathbf{77.2}_{\pm0.6}$
    & $\mathbf{79.6}_{\pm1.1}$
    & $\mathbf{0.840}_{\pm0.006}$
    & $\mathbf{75.8}_{\pm0.3}$
    & $88.3_{\pm0.6}$
    & $125.7_{\pm0.8}$ \\
LocAgent
    & $53.8_{\pm0.5}$
    & $74.8_{\pm0.3}$
    & $79.1_{\pm0.3}$
    & $0.791_{\pm0.003}$
    & $73.7_{\pm0.6}$
    & $\mathbf{94.4}_{\pm0.7}$
    & $1292.7_{\pm4.4}$ \\
RepoSearcher
    & $56.2_{\pm0.4}$
    & $74.4_{\pm1.0}$
    & $74.4_{\pm1.0}$
    & $0.796_{\pm0.004}$
    & $\mathbf{75.8}_{\pm0.8}$
    & $88.6_{\pm0.8}$
    & $\mathbf{72.1}_{\pm2.9}$ \\
Agentless
    & $39.6_{\pm1.0}$
    & $61.6_{\pm1.9}$
    & $63.1_{\pm1.6}$
    & $0.613_{\pm0.022}$
    & $70.0_{\pm1.2}$
    & $81.9_{\pm1.1}$
    & -- \\
\bottomrule
\end{tabular}
\end{table}

\paragraph{Accuracy gains and token tradeoffs.}
On SWE Bench Verified, CuraLoc achieves the highest
mean Func@1 at 59.3\%, versus LocAgent's 53.8\%,
RepoSearcher's 56.2\%, and Agentless's 39.6\%
(Table~\ref{tab:method-comparison}).
Its 125.7K tokens include 43.1K for summary generation.
Against LocAgent, it uses 90.3\% fewer tokens;
multiplicity-adjusted joint bounds support Pareto
dominance, with lower bounds of 1.7 accuracy points
and 1.08 million tokens saved per task.
Both advantages remain supported under repository resampling.

Against RepoSearcher, CuraLoc also raises Func@5 from
74.4\% to 77.2\%, at $1.74\times$ its recorded tokens.
Pointwise task-bootstrap 95\% intervals for the
Func@1 and Func@5 gains are $[0.95,5.43]$ and
$[0.66,4.77]$ points.
Against Agentless, Func@10 rises from 63.1\% to 79.6\%.
LocAgent retains higher File@10.

\paragraph{The advantage extends beyond configuration-selection tasks.}
On the 370 SWE Bench Verified tasks absent from
SWE Bench Lite, Func@1 is 55.5\% for CuraLoc,
52.5\% for RepoSearcher, 51.0\% for LocAgent,
and 37.0\% for Agentless.
The CuraLoc--RepoSearcher difference has a pointwise
task-bootstrap 95\% interval of $[0.45,5.59]$ points.
CuraLoc uses approximately one-tenth LocAgent's tokens,
with family-adjusted task-bootstrap support for both
accuracy and token advantages.
Full subset results appear in
Appendix~\ref{app:verified-cohorts}.

\begin{table}[!htbp]
\centering
\caption{GPT-5.2 localization on SWE Bench Lite ($n=274$), using
the conventions of Table~\ref{tab:method-comparison}.
SWE Bench Lite was used for configuration selection.
RepoSearcher returns at most five symbols, so its
Func@10 equals Func@5.}
\small
\setlength{\tabcolsep}{3pt}
\begin{tabular}{lrrrrrrr}
\toprule
Method & Func@1 & Func@5 & Func@10 & MRR
       & File@1 & File@10 & Tokens (K) \\
\midrule
CuraLoc
    & $\mathbf{64.4}_{\pm1.6}$
    & $\mathbf{83.8}_{\pm1.6}$
    & $\mathbf{87.1}_{\pm0.2}$
    & $\mathbf{0.798}_{\pm0.009}$
    & $\mathbf{85.9}_{\pm0.9}$
    & $93.6_{\pm0.2}$
    & $124.4_{\pm0.6}$ \\
LocAgent
    & $58.5_{\pm1.1}$
    & $82.2_{\pm0.4}$
    & $85.5_{\pm1.1}$
    & $0.755_{\pm0.006}$
    & $82.0_{\pm1.4}$
    & $\mathbf{97.0}_{\pm0.2}$
    & $1457.4_{\pm21.4}$ \\
RepoSearcher
    & $61.2_{\pm1.1}$
    & $78.3_{\pm1.7}$
    & $78.3_{\pm1.7}$
    & $0.759_{\pm0.011}$
    & $84.1_{\pm1.3}$
    & $93.1_{\pm0.7}$
    & $80.3_{\pm3.8}$ \\
Agentless
    & $46.8_{\pm1.5}$
    & $69.8_{\pm1.2}$
    & $70.9_{\pm0.8}$
    & $0.625_{\pm0.008}$
    & $79.9_{\pm0.6}$
    & $89.7_{\pm1.1}$
    & $\mathbf{72.7}_{\pm2.8}$ \\
\bottomrule
\end{tabular}
\label{tab:lite-results}
\end{table}

\begin{figure}[!htbp]
\centering
\includegraphics[width=\linewidth]{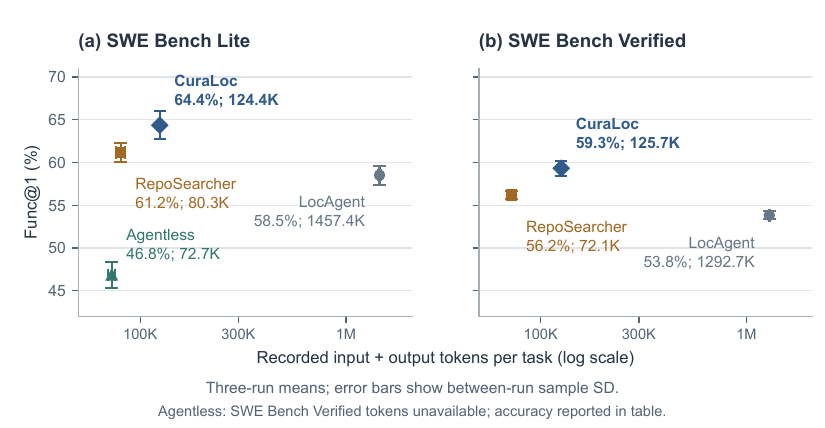}
\caption{Func@1 versus recorded tokens on
(a) SWE Bench Lite and (b) SWE Bench Verified.
Points are three-run means; error bars show sample
standard deviations. The token axis is logarithmic.
Agentless is omitted from panel (b) because its exact
token total is unavailable.}
\label{fig:verified-results}
\end{figure}

\paragraph{Higher accuracy at greater computation than cheaper alternatives.}
On SWE Bench Lite, CuraLoc achieves the highest observed mean
at every function cutoff and in MRR
(Table~\ref{tab:lite-results}).
Against Agentless, Func@1 rises from 46.8\% to 64.4\%
at $1.71\times$ its recorded tokens.
Against RepoSearcher, Func@5 rises from 78.3\% to 83.8\%
at $1.55\times$ its tokens; the paired 95\% interval
for this accuracy difference is $[2.6,8.5]$ points.
The Func@1 difference from RepoSearcher remains
statistically unresolved. Figure~\ref{fig:verified-results} therefore distinguishes two outcomes: CuraLoc improves both objectives relative
to LocAgent, but requires additional tokens for its
higher observed accuracy relative to the lower-token
alternatives on SWE Bench Lite.

\begin{figure}[!htbp]
\centering
\includegraphics[width=.85\linewidth]{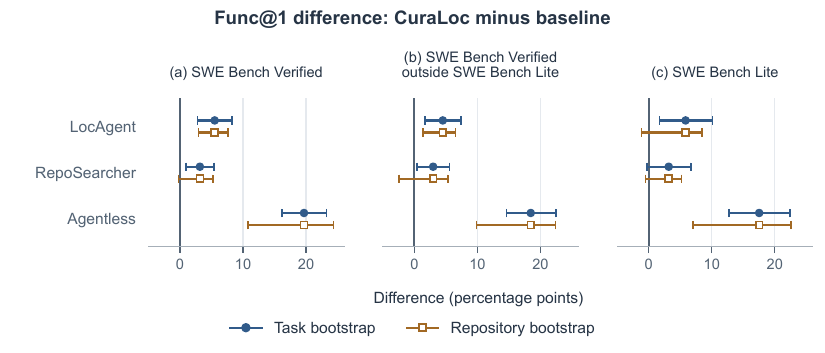}
\caption{CuraLoc-minus-baseline Func@1 differences with
pointwise 95\% intervals from paired task and
whole-repository resampling. Positive values favor
CuraLoc. These intervals are unadjusted; SWE Bench Verified
outside SWE Bench Lite is a subset of SWE Bench Verified.}
\label{fig:paired-effects-main}
\end{figure}

On SWE Bench Verified and its subset absent from
SWE Bench Lite, pointwise Func@1 intervals favor CuraLoc
over LocAgent and Agentless under both resampling
procedures. The RepoSearcher differences are positive
under task resampling but unresolved under repository
resampling. All comparisons appear in
Figure~\ref{fig:paired-effects} in the appendix.

\subsection{Code Selection and Inspection}
\label{sec:ablations}

\paragraph{Broad assessment, selective source retrieval.}
Across the three SWE Bench Verified runs, the per-run medians
are 320.5--335 candidate descriptions, seven starting
symbols, and 22--25 symbols selected for source retrieval,
including neighbors. These counts precede input-budget
truncation; post-budget source-block lists were not
recorded.

\paragraph{Component comparisons.}
Each intervention is evaluated once on SWE Bench Lite
against the same saved development run, which achieves
67.2\% Func@1 (Table~\ref{tab:ablations}).
We report paired differences from this reference,
reusing unchanged upstream artifacts as specified in
Appendix~\ref{app:ablation-details}.
The reference is separate from the three-run cold
evaluation reported in Table~\ref{tab:lite-results}.

\begin{table}[!htbp]
\centering
\caption{Single-run SWE Bench Lite comparisons ($n=274$) against
one fixed reference. Interventions are separate;
the selection-stage ranking is an offline readout.}
\small
\setlength{\tabcolsep}{6pt}
\begin{tabular}{lrrr}
\toprule
Configuration & Func@1 & Func@10 & MRR \\
\midrule
Full CuraLoc (reference)    & 67.2 & 87.6 & 0.823 \\
Selection-stage readout     & 55.8 & 83.2 & 0.739 \\
Source-order selection     & 42.3 & 55.1 & 0.518 \\
Without summaries          & 64.6 & 86.1 & 0.807 \\
Starting symbols only      & 67.2 & 83.9 & 0.822 \\
Relevance-guided formulation & 63.1 & 82.8 & 0.775 \\
\bottomrule
\end{tabular}
\label{tab:ablations}
\end{table}

\paragraph{Selecting useful symbols matters within the correct files.}
Reference file scoping includes all gold files within
ten predictions on 98.2\% of tasks.
Replacing model-guided selection and its annotations
with selection by file rank and source-code order
preserves the candidate pool and evidence-symbol count,
but lowers Func@1 from 67.2\% to 42.3\%.
The selected evidence includes all reference edit
symbols on 54.0\% of tasks, versus 90.5\% for CuraLoc.
The difference therefore extends beyond finding
the correct files or selecting more symbols.

\paragraph{Source-based ranking improves description-based predictions.}
The ranking extracted from S2 achieves 55.8\% Func@1.
Holding its RouteMap fixed, a final ranking call using
source code for the starting symbols reaches 67.2\%,
without neighborhood expansion.
This comparison measures source inspection and the
additional ranking decision together.

\paragraph{Summary removal.}
Removing behavioral summaries and recomputing S2 and
S3 yields 64.6\% Func@1 and 86.1\% Func@10.
At Func@1, the intervention gains six tasks and loses
thirteen relative to the reference. Recorded S2+S3
tokens decrease by 27.0\%; these stage totals exclude
S1 and summary generation.

\paragraph{Expansion supplies missing edit locations.}
Adding related code raises Func@10 from 83.9\% to 87.6\%,
with 11 paired gains and one loss.
All 274 comparisons use byte-identical S2 artifacts.
Reconstructing the reference selection with its frozen
retrieval code shows that every gained task adds
reference edit symbols absent from the starting selection.
Five gains involve single-symbol edits and six involve
multi-symbol edits. Among the 35 multi-symbol tasks,
complete recovery rises from 19 to 24 tasks.
Expansion improves complete edit-set recovery at
$3.5\times$ the final-ranking tokens, with unchanged
aggregate Func@1 in this comparison.

\paragraph{Guidance formulation affects localization.}
Replacing the responsibility-oriented perspectives,
roles, and guidance across all three stages with
relevance-oriented alternatives yields 63.1\% Func@1
and 82.8\% Func@10, below the reference's 67.2\% and 87.6\%.
Both formulations retain explanations, hypotheses,
the same models, summarization policy, structural
operations, and configured limits, while selecting
their own files and evidence.
Appendix~\ref{app:ablation-details} specifies the
interventions and artifact provenance.

\paragraph{Where localization still fails.}
Across three SWE Bench Verified runs, 93.9\% of CuraLoc's
Func@10 failures involve a reference edit symbol
that was not selected for S3. When all reference
symbols are selected, the final ranking recovers
the complete set in 98.4\% of executions.
LocAgent often encounters the required code but
does not return the complete edit set: all reference
implementations appear in its tool outputs in
182 of 284 failed executions.
On tasks with complete coverage under both methods'
respective definitions, CuraLoc recovers the complete
edit set on 22--24 more tasks per run
(Table~\ref{tab:joint-coverage-recovery}).
Appendix~\ref{app:failure-analysis} details the
method-specific coverage definitions and paired results.

\subsection{Evidence Breadth and Source Depth}
\label{sec:hyperparameters}

\begin{figure}[!htbp]
\centering
\includegraphics[width=\linewidth]{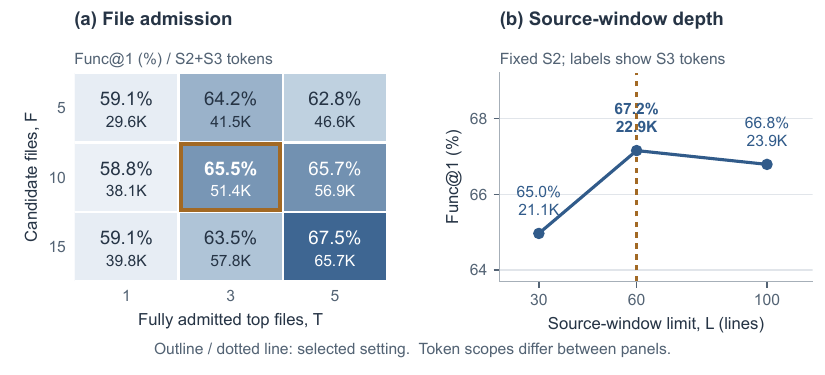}
\caption{Evidence-budget sensitivity on SWE Bench Lite ($n=274$).
(a) Func@1 and mean S2+S3 tokens across file settings;
the outlined setting averages three development runs,
while alternatives use one run each.
(b) Source depth with S2 fixed to one reference run;
labels show Func@1 and S3 tokens.
Both panels exclude S1 and summary-generation tokens.}
\label{fig:hp-sweep}
\end{figure}

We vary candidate-file count $F$, fully included
top-file count $T$, and source-window limit $L$
on SWE Bench Lite (Figure~\ref{fig:hp-sweep}).

\paragraph{Broader admission trades tokens for accuracy.}
We retain $F=10,T=3$ as the selected setting.
Increasing to $F=15,T=5$ raises development Func@1
from 65.5\% to 67.5\% at 27.9\% more S2+S3 tokens.
The highest Func@10, 88.0\%, occurs at $F=10,T=5$.
No setting maximizes accuracy at every cutoff.

\paragraph{Longer source windows do not consistently improve ranking.}
With the S2 output fixed, $L=60$ achieves 67.2\% Func@1,
compared with 65.0\% at $L=30$ and 66.8\% at $L=100$.
Func@5 and Func@10 remain unchanged.
Increasing the window from 60 to 100 lines adds 4.2\%
to final-ranking tokens without improving accuracy
in this comparison. Full development results appear
in Appendix~\ref{app:hp-sweep}.

\subsection{Additional-Backbone Evaluation}
\label{sec:backbone}

We evaluate all four methods with GPT-6 Astra in one
run each on SWE Bench Verified, reporting localization
accuracy and recorded tokens
(Table~\ref{tab:gpt6-results}).
\begin{table}[!htbp]
\centering
\caption{Single-run GPT-6 Astra localization on SWE Bench
Verified ($n=454$). Tokens are in thousands per task
and include all contributing model calls, including
CuraLoc's GPT-5.4-mini summaries. Bold marks the best
score or lowest token total. RepoSearcher's Func@10
equals Func@5; finalized empty rankings count as misses.}
\label{tab:gpt6-results}
\small
\setlength{\tabcolsep}{2.5pt}
\begin{tabular}{lrrrrrrrr}
\toprule
Method & Func@1 & Func@3 & Func@5 & Func@10
       & MRR & File@1 & File@10 & Tokens (K) \\
\midrule
CuraLoc
    & \textbf{67.0} & \textbf{84.4} & \textbf{87.4}
    & \textbf{89.2} & \textbf{0.935}
    & \textbf{84.1} & 94.5 & 138.9 \\
LocAgent
    & 63.2 & 80.0 & 82.2 & 85.9
    & 0.896 & 81.1 & \textbf{96.9} & 95.3 \\
RepoSearcher
    & 66.7 & 82.8 & \textbf{87.4} & 87.4
    & 0.926 & 83.9 & 94.1 & 105.0 \\
Agentless
    & 59.7 & 76.2 & 78.4 & 79.7
    & 0.846 & 82.2 & 90.3 & \textbf{28.2} \\
\bottomrule
\end{tabular}
\end{table}

CuraLoc exceeds LocAgent and Agentless at every reported
function cutoff and in MRR. Against RepoSearcher,
Func@1 differs by one task (304 versus 303), while
Func@5 is identical at 87.4\%.
On the 370 SWE Bench Verified tasks absent from
SWE Bench Lite, Func@1 is 62.4\%, 61.9\%, 59.5\%,
and 56.8\% for CuraLoc, RepoSearcher, LocAgent, and
Agentless, respectively; Func@5 is 84.9\%, 84.9\%,
80.3\%, and 76.2\%. CuraLoc uses 89.2K GPT-6 tokens per task versus
LocAgent's 95.3K. Its GPT-5.4-mini summaries add
49.7K tokens, bringing the total to 138.9K.
Thus, higher accuracy uses fewer primary-model
tokens but more total tokens than LocAgent.

\section{Discussion}
\label{sec:discussion}

\paragraph{Adaptive inspection under a fixed decision schedule.}
The comparisons show that CuraLoc's fixed decision
schedule supports more than a reduction in token use:
it exceeds all three baselines in Func@1 on SWE Bench
Verified with GPT-5.2 and remains competitive with
RepoSearcher under GPT-6.
The methodological contribution is planning source
inspection from compact behavioral and structural
descriptions. Compact descriptions
support joint choices of starting symbols and retrieval
actions; selected implementations support the final
edit decision. This allows the inspected code to adapt
without extending the sequence of model decisions.
The accuracy--token tradeoff depends on the backbone:
CuraLoc improves both objectives over LocAgent with
GPT-5.2, while its higher accuracy with GPT-6 Astra
uses more total tokens when summary generation
is included.

\paragraph{The value of additional code depends on the decision.}
The component experiments distinguish selecting useful
code from ranking edit locations within it
(\S\ref{sec:ablations}).
Model-guided selection improves coverage at the same
symbol count, and source inspection with a final ranking
improves the description-stage prediction.
Expansion serves a different purpose: it recovers
additional edit locations without changing aggregate
Func@1 in the paired diagnostic.
The source-depth sweep likewise shows that longer
snippets do not consistently improve ranking.
The failure analysis adds a complementary observation:
encountering the reference implementations does not
ensure their recovery in the final ranking.
Together, these findings motivate treating source
selection and final comparison as distinct design
decisions, rather than measuring investigation only
by how much relevant code it reaches.

\paragraph{Limitations.}
Our evaluation focuses on pre-existing Python symbols
in SWE Bench. The SWE Bench Verified tasks absent from SWE Bench Lite
exclude configuration-selection instances, but remain
within the same benchmark and repository collection;
transfer to other languages and repositories remains
to be established. Reference patches identify one
observed fix, although other correct fixes may modify
different code. Our results therefore measure recovery
of reference edit locations rather than every valid
repair strategy. Establishing whether the resulting
rankings improve repair success requires a downstream
evaluation with the repair procedure held fixed.

\section{Conclusion}
\label{sec:conclusion}

We introduced CuraLoc to test whether repository-level code inspection can be planned before source retrieval rather than directed through repeated model-controlled navigation. CuraLoc selects candidate files, plans starting symbols and relationship-specific retrieval actions from compact symbol descriptions, retrieves the requested source deterministically, and ranks edit locations in a final model decision. On SWE Bench Verified with GPT-5.2, CuraLoc improves Func@1 over all three baselines while using 90.3\%
fewer recorded tokens than LocAgent. With GPT-6 Astra,
it remains competitive with RepoSearcher and outperforms
LocAgent and Agentless at the reported function cutoffs. These results show that issue-specific inspection can remain adaptive without requiring an iterative sequence of model-directed retrieval steps. Future work should test whether this tradeoff holds across repositories, languages, and downstream repair settings.

\subsection*{AI use statement}

We used generative AI tools to assist with method and
prompt refinement, including the role vocabulary;
experiment design; implementation of methods and
analysis code; and interpretation of results.
We also used these tools for literature discovery,
figure preparation, and drafting and editing the
manuscript. Verification included automated tests,
reconciliation of reported metrics and token totals
with saved experimental artifacts, and checks of cited
sources. The authors reviewed the AI-assisted work
and take responsibility for the final text, claims,
code, and results.

\subsection*{Reproducibility statement}

Section~\ref{sec:method} and
Appendix~\ref{app:method-details} specify the method,
representations, and retrieval policies.
Appendix~\ref{app:eval-details} documents the cohorts,
normalization, model configurations, and repetition
and cache policies.
Appendix~\ref{app:statistics} describes the statistical
analysis; Appendices~\ref{app:timing-details},
\ref{app:ablation-details}, and \ref{app:hp-sweep}
document token accounting, component interventions,
and configuration sweeps.

\bibliographystyle{plainnat}
\bibliography{references}

\appendix
\section*{Appendix}
\section{Method Implementation Details}
\label{app:method-details}

This section specifies the representations, deterministic policies,
and output handling underlying Algorithm~\ref{alg:CuraLoc}. Graph-role
definitions, edge resolution, the issue-conditioned role taxonomy,
and a capsule example follow in Appendices~\ref{app:graph-roles},
\ref{app:edge-resolution}, \ref{app:role-taxonomy}, and
\ref{app:capsule-example}, respectively.

\subsection{File Scoping and Supplementation}
\label{app:file-scoping}

Stage~1 deterministically constructs two repository representations
and selects the one with the smaller token count. The \emph{file
skeleton} lists top-level classes and functions; the \emph{class map}
additionally lists class methods. Python test files appear only as
paths in both views, retaining their candidacy without serializing
their symbol inventories.

For each perspective $p$, the model proposes 6--10 files with a brief
rationale and a consumer probability $q_{p,f}\in[0,1]$ per file.
Values $q_{p,f}\geq0.5$ indicate consumer-facing locations; lower
values indicate mechanism locations. The perspective vote count is
$v(f)=\sum_p\mathbb{I}[f\text{ is proposed under }p]$. Files are
ranked by decreasing $v(f)$ and then first appearance, and the union
is truncated to at most 15 files. The output retains perspective
provenance, votes, probabilities, and rationales. Ownership, runtime
participation, and enforcement may reside in different files, allowing
multiple causal explanations to survive this initial scoping step.

When Stage~1 returns fewer files than the downstream policy requires,
its seeds may be supplemented by structurally neighboring files.
File relations come from imports, calls, and inheritance. A neighbor
receives weight 3 for inheritance, 2 for import, and 1 for call
connectivity, plus one point when supported by multiple relation types.
Expansion adds at most two neighbors per seed and eight overall, with
at most 25 total graph nodes. Original Stage~1 seeds precede expanded
files; symbol admission still applies the leading-$F$ file limit.

\subsection{Structural Inventory and Capsules}
\label{app:structural-details}

The semantic AST includes files, classes, methods, top-level functions,
available module variables, imports, inheritance, calls, and source
locations. Stable integer SIDs bind model outputs and graph operations
to exact symbols without repeatedly serializing file-qualified names.
Directed relations are \texttt{member\_of} (class to method),
\texttt{inherits} (subclass to resolved base), \texttt{overrides}
(subclass method to base method), \texttt{calls} (caller to callee),
and its reciprocal \texttt{called\_by}. Call resolution prioritizes
self-qualified, same-file qualified and bare, unique cross-file
qualified, and module-qualified path-suffix matches; ambiguous
cross-file bare calls are omitted (Appendix~\ref{app:edge-resolution}).
Import adjacency is symmetric at file level and affects admission,
not Stage~3 navigation.

Topology-derived roles are \emph{executor}, \emph{leaf}, \emph{hub},
\emph{entry}, \emph{dispatcher}, \emph{definer}, and \emph{overrider}.
They describe graph position rather than issue-conditioned
responsibility. Their thresholds and assignment precedence are given
in Appendix~\ref{app:graph-roles}.

\paragraph{Capsule fields.}
\label{sec:symbol-capsules}
Each admitted symbol's capsule contains its SID, qualified name,
signature, kind, file rank, top/context-file status, Stage~1
consumer/mechanism evidence, deterministic graph role, local degree,
bounded call/caller/inheritance/override adjacency, and behavioral
summary when available. Prompt adjacency is capped to limit tokens;
the complete resolved graph remains available for deterministic
expansion. The primary Stage~2 call receives the issue and capsules,
not raw source bodies. Appendix~\ref{app:capsule-example} illustrates
the SID-bound format.

\subsection{Issue-Independent Behavioral Summaries}
\label{sec:summaries}

Summaries are constructed before Stage~2, independently of the issue,
so unchanged code can reuse descriptions across different tasks.
Summary text is selected in order from (1) the first paragraph of the
symbol's opening docstring, if at most 400 characters; (2) an existing
repository-cache entry; or (3) the auxiliary model. Generated summaries
are limited to 400 characters.

The summarizer reads at most 100 source lines per admitted symbol,
without extra context or a prepended class header; longer symbols
receive an explicit truncation marker. Cache keys hash the
summarization prompt version, symbol kind, and source text, permitting
reuse only when the summarized code and prompt semantics are unchanged.
Uncached symbols are batched in groups of at most 25 symbols and
24,000 source characters. Missing outputs are retried in progressively
smaller batches. At least 95\% of symbols requiring model-generated
descriptions must receive summaries; otherwise the task fails rather
than continuing with substantially incomplete representations.
Section~\ref{sec:ablations} evaluates summary removal.

\subsection{RouteMap Validation}
\label{app:route-validation}

The Stage~2 prompt requests relevant symbols with relevance scores,
one-sentence rationales, and one of eight issue-conditioned roles
(Appendix~\ref{app:role-taxonomy}); five to eight ordered starting
points with a hypothesis and action each; and two to four global
causal hypotheses describing possible causes and supporting source
evidence. The requested starting-point count is a prompt instruction rather
than a hard parser constraint: a nonempty valid list outside the
requested range is accepted and recorded diagnostically. These semantic roles are distinct from deterministic graph
roles. Global hypotheses enter Stage~3 as prior diagnostic context,
without separate model calls to validate them.

Emitted SIDs absent from the admitted inventory are discarded and
relevance scores are clamped to $[0,1]$. Malformed RouteMaps,
including those with no valid starting points, are rejected. A validated
RouteMap stores the admitted inventory, semantic annotations, starting
points, actions, hypotheses, navigation graph, summaries, and
construction diagnostics.

\subsection{Neighborhood and Prompt Ordering}
\label{app:evidence-order}

For \texttt{expand\_local}, same-class neighbors are ordered by their
Stage~2 status and relevance, direct structural connectivity to the
starting point, and source position. Other actions use deterministic
graph adjacency order. After the starting points, neighbors enter
round-robin across starting points, rather than exhausting one large
class or high-degree neighborhood. Duplicates retain their first
occurrence. Expansion is one hop, never recursively follows new
neighbors, and cannot select symbols outside the admitted inventory.
Evidence breadth depends on starting-point count and local connectivity,
not a fixed symbol-count limit.

The prompt separates \emph{assessed candidates} (starting points and
other symbols assigned roles or relevance by Stage~2) from
\emph{adjacent evidence} (previously unassessed symbols reached by
expansion). Assessed candidates are ordered by Stage~2 shortlist
position when available, otherwise by relevance. Each block may
contain semantic role, relevance, rationale, behavioral summary,
starting-point hypothesis, and source. Global hypotheses and a compact
names-only inventory of the highest-ranked relevant symbols are also
included.
Each RouteMap starting point specifies one investigation action.
The action determines the eligible one-hop neighborhood:

\begin{description}[style=nextline,leftmargin=0pt]
    \item[\texttt{expand\_local}]
    Other methods in the starting symbol's class.

    \item[\texttt{follow\_calls}]
    Callees connected by outgoing \texttt{calls} edges.

    \item[\texttt{follow\_called\_by}]
    Callers identified by the stored caller relationships.

    \item[\texttt{check\_override}]
    Related base and overriding implementations.
\end{description}

All selected neighbors must belong to the admitted symbol inventory.
Starting points and actions determine evidence collection; the
accompanying hypotheses provide diagnostic context for the final
ranking rather than triggering separate investigations.

\subsection{Source Rendering and Fallbacks}
\label{sec:source-evidence}
Source is retrieved from the task's base-commit checkout. The primary
source slice is capped at $L=60$ lines, including the context selected
by the three-line context setting. For methods, a separately prepended
class header can add up to 20 lines. Thus $L$ bounds the primary slice,
not the entire rendered evidence block.

If AST line metadata is missing, CuraLoc searches the file for the symbol
definition. If no definition is found, it includes the first 30 file
lines and records a warning. Whenever a symbol extends beyond the
displayed slice, the prompt explicitly marks the source as truncated.

\subsection{Budget Enforcement and Output Handling}
\label{app:ranking-details}

Let $\mathcal{B}=(b_1,\ldots,b_N)$ be evidence blocks in directed
priority order. If the complete input exceeds the budget, CuraLoc retains
the longest complete prefix $(b_1,\ldots,b_j)$ that fits. It stops at
the first non-fitting block without backfilling shorter, lower-priority
blocks. The issue, prior diagnosis, source-window policy, and ranking
instructions remain unchanged. Truncation follows the established
starting-point and round-robin priority rather than using gold labels,
new model scores, or task identity to choose omissions. The main-run artifacts retain the symbols selected
for S3 and the final ranking. They do not separately
record which source snippets, if any, were removed
to fit the input limit. The final runtime uses an estimated input budget of 272,000 tokens.
The estimate covers the serialized system and user messages and
includes a 1,024-token safety margin. Prompts within the budget are
unchanged.
The ranking model receives the issue, prior diagnosis, and source
evidence in one call and is asked for up to ten edit symbols,
restricted to symbols represented in the prompt. Exact file-qualified
duplicates are removed without reordering; fewer than ten valid
predictions may remain. With one Stage~3 sample, positional Borda
normalization preserves that list's order while attaching normalized
rank scores and appearance metadata. The optional second-phase
tie-breaker and iterative hypothesis validation are disabled. Finally,
predictions are resolved against the semantic AST and enriched with
symbol kinds and source ranges for evaluation.

\section{Graph Role Definitions}
\label{app:graph-roles}

Each symbol in the classification set receives a topology-derived graph role 
computed from its position in the call and override graph. Roles are assigned 
by the following priority rules (later overrides earlier):

\begin{itemize}[nosep,leftmargin=*]
    \item \emph{executor}: default (no special topological property).
    \item \emph{leaf}: out-degree 0, in-degree $\geq 1$.
  \item \emph{hub}: degree $\geq$ 90th percentile within its file, or $\geq 8$.
  \item \emph{entry}: has cross-file callers but no in-file callers.
    \item \emph{dispatcher}: out-degree $\geq 3$ with the sum of distinct
        callee files and file-qualified callee classes $\geq 2$.
    \item \emph{definer}: is the target of an override relation.
  \item \emph{overrider}: has outgoing override edges.
\end{itemize}

Degree counts include call and override relations. The per-file hub
threshold uses the sorted degree at zero-based index
$\lfloor 0.9n \rfloor$ for files containing $n\geq 5$ symbols;
smaller files use a threshold of 8.

\section{Edge Resolution Confidence}
\label{app:edge-resolution}

Call resolution uses repository structure to avoid speculative links:

\begin{enumerate}[nosep,leftmargin=*]
  \item \emph{self-qualified}: \texttt{self.method()} resolved within the 
    same class.
  \item \emph{qualified same-file}: \texttt{Class.method()} found in the 
    same file.
  \item \emph{bare-name same-file}: unqualified function name matched in 
    the same file.
  \item \emph{qualified cross-file}: a unique qualified name or
    module-qualified path suffix resolves a symbol in another candidate file.
\end{enumerate}

Ambiguous cross-file bare calls are omitted. The resulting resolved
call graph supplies both outgoing \texttt{calls} and reciprocal
\texttt{called\_by} relations for directed evidence construction.

\section{LLM Role Taxonomy}
\label{app:role-taxonomy}

We developed the role vocabulary through author--LLM design
discussions, refining earlier role schemes using examples of
localization failures. The resulting eight roles describe
issue-conditioned relationships that can inform evidence
selection and edit-location assessment. Stage~2 assigns these
labels using the issue description and symbol representations:

\begin{itemize}[nosep,leftmargin=*]
  \item \textsc{entry\_point}: entry to the subsystem relevant to the issue.
  \item \textsc{dispatcher}: routes execution to multiple targets.
  \item \textsc{behavior\_definer}: implements the core logic relevant to 
    the bug.
  \item \textsc{invariant\_enforcer}: enforces a contract or validation rule.
  \item \textsc{state\_provider}: provides or manages state consumed by 
    other symbols.
  \item \textsc{concrete\_owner}: concrete implementation of an abstract 
    contract.
  \item \textsc{base\_contract}: abstract base class or interface defining 
    the contract.
  \item \textsc{companion}: sibling method that typically changes alongside 
    the primary fix target.
\end{itemize}

\section{Capsule Format Example}
\label{app:capsule-example}

This excerpt shows recorded S2 inventory fields for
\texttt{django\_\_django-11001} from the fixed development
reference. Original symbol identifiers, SIDs, signatures,
and field values are preserved; layout and line wrapping
are reformatted. It illustrates the stored representation,
not the exact serialized model input.

\begin{verbatim}
[SID=435] django/db/models/query.py:QuerySet.filter
signature: (self, *args, **kwargs)
summary: Return a new QuerySet instance with the args ANDed to the
  existing set.
calls: django/db/models/query.py:QuerySet._filter_or_exclude

[SID=431] django/db/models/query.py:QuerySet.exclude
signature: (self, *args, **kwargs)
summary: Return a new QuerySet instance with
  NOT (args) ANDed to the existing set.
calls: django/db/models/query.py:QuerySet._filter_or_exclude

[SID=401] django/db/models/query.py:QuerySet._filter_or_exclude
signature: (self, negate, *args, **kwargs)
calls: django/db/models/query.py:QuerySet._chain
\end{verbatim}

\noindent The summaries distinguish the two public methods,
while the call relationships identify their shared helper.
Displayed calls agree with the saved graph and base-commit
source. Caller lists, the helper's summary, and other
fields are omitted; recorded adjacency lists are capped
at five entries.

\section{Evaluation Protocol Details}
\label{app:eval-details}

\paragraph{Tasks and reference labels.}
Each method receives an issue description and its pre-fix repository
snapshot. Reference patches supply the evaluation targets and are
not provided during localization. Symbol-level evaluation uses the
274 SWE Bench Lite and 454 SWE Bench Verified instances with scoreable pre-existing
function- or class-level edit labels. The corresponding gold files
are the files containing those symbols, rather than every file
changed by the reference patch. All methods use the same task sets.
Of the scoreable SWE Bench Verified instances, 84 occur in SWE Bench Lite and 370 do not.

\paragraph{Normalization and matching.}
Gold labels and predictions use repository-relative,
file-qualified symbol identifiers of the form
\texttt{file.py:Symbol}. The canonical normalizer strips surrounding
whitespace, Markdown asterisks and backticks, argument signatures,
and trailing constructor suffixes. Thus,
\texttt{file.py:Class.\_\_init\_\_} maps to
\texttt{file.py:Class}. The same normalization is applied to
predictions and gold labels. Already-normalized exports are not
normalized a second time. Duplicate predictions are removed while
preserving their first occurrence. Matching is exact; no fuzzy,
substring, or partial-symbol credit is awarded.

\paragraph{Localization metrics.}
Func@$k$ counts an instance as correct only when all normalized
gold symbols appear among its first $k$ distinct predictions.
File@$k$ applies the same requirement to the first $k$ distinct
files derived from that symbol ranking, preserving first occurrence.
It does not use a separately generated file ranking. Native
file-scoping diagnostics are identified separately.

Mean reciprocal rank (MRR) averages the reciprocal rank of the
first matching gold symbol, assigning zero when none is predicted.
Consequently, partial recovery of a multi-symbol edit set can
receive MRR credit while failing Func@$k$. Empty finalized rankings
and finalized execution failures count as misses; unresolved runs
are not scored over a reduced successful-task subset.

\paragraph{GPT-5.2 configurations.}
CuraLoc uses GPT-5.2 at temperature~0.3 with $F=10$, $T=3$,
$L=60$, symmetric import-based admission, directed neighborhood
expansion, and one final ranking sample. Behavioral summaries use
GPT-5.4-mini. The tie-breaking and iterative hypothesis-validation
variants are disabled.

Baselines retain their inference configurations with GPT-5.2 as
the primary model. LocAgent uses temperature~1.0 and combines
two samples through reciprocal-rank fusion; line-to-symbol
resolution is restricted to the originating file.
RepoSearcher and Agentless use the model-default temperature
in the evaluated API configuration. RepoSearcher returns at most
five symbol predictions, so Func@10 equals Func@5.
Agentless uses four final localization samples. These are
same-primary-model comparisons, not matched sampling budgets.

\paragraph{Repetitions and cache policy.}
Main GPT-5.2 tables aggregate three completed runs for each
included method--dataset combination. Metrics are computed
separately per run, then averaged; reported standard deviations
are sample standard deviations across the three run scores.
Predictions are not ensembled across benchmark repetitions.

Main CuraLoc runs disable summary-cache reuse across tasks and runs.
Resumption may reuse completed stages from the same run, with
their original contributing requests counted once. Development
sweeps and ablations retain their designated reference artifacts
and are reported separately from these cold-run comparisons.

\paragraph{Additional-backbone comparison.}
The GPT-6 Astra evaluation includes one SWE Bench Verified
run per method. CuraLoc and LocAgent use temperature~1.0;
RepoSearcher and Agentless use the provider-default
setting with the temperature parameter omitted.
CuraLoc uses one final ranking sample and GPT-5.4-mini
summaries; LocAgent combines two trajectories; Agentless
aggregates four final samples. RepoSearcher uses one
sample per call and returns at most five symbols.
RepoSearcher and Agentless each have six finalized
empty rankings among the 454 scoreable tasks, counted
as misses. All methods use the same normalization,
metrics, and cohorts. We report single-run accuracy and retained input-plus-output tokens over the same 454 tasks, counting all contributing
stages and samples, including CuraLoc's summary generation.
\section{Statistical Analysis}
\label{app:statistics}

\paragraph{Run variability and paired differences.}
For each GPT-5.2 method, we compute localization metrics
and, where usage is complete, average recorded tokens
per task separately in each of three runs.
The single-run GPT-6 Astra results are descriptive
and excluded from these bootstrap analyses. We report the mean
and sample standard deviation of these run-level measurements.
Standard deviations describe variation across executions; they are
not confidence intervals.

For paired comparisons, we first average each task's outcomes across
the three runs within each method. We then compute the difference
between CuraLoc and the baseline on that task. Averaging these differences
over the cohort gives the difference between the methods' reported
three-run means. Run numbers are not treated as matched random seeds,
and repeated executions of a task are not treated as independent tasks.

\paragraph{Paired task bootstrap.}
We draw 50,000 samples of task IDs with replacement, each containing
the original cohort's number of tasks. Each selected task retains
all three recorded executions, and the same sampled IDs are used
for every method, metric, and token calculation. Repeatedly selected
tasks contribute with multiplicity.

For each resample, we recompute the method differences. Pointwise
95\% percentile intervals use the 2.5th and 97.5th percentiles
of the resulting distributions. Accuracy differences are reported
in percentage points; MRR differences remain on the 0--1 scale.
These intervals describe uncertainty across benchmark instances
conditional on the recorded executions, rather than the full
variability of future model runs.

\paragraph{Repository sensitivity.}
Each evaluated cohort contains 12 repositories. To examine dependence
on repository composition, we also draw 50,000 samples of whole
repositories with replacement. A selected repository contributes
all its tasks and their three executions; repeated selections retain
that multiplicity. Metrics remain task-weighted: we divide the
summed outcomes by the number of sampled tasks, rather than averaging
repository scores equally. We additionally recompute effects after
omitting each repository in turn.

\begin{figure}[!htbp]
\centering
\includegraphics[width=\linewidth]{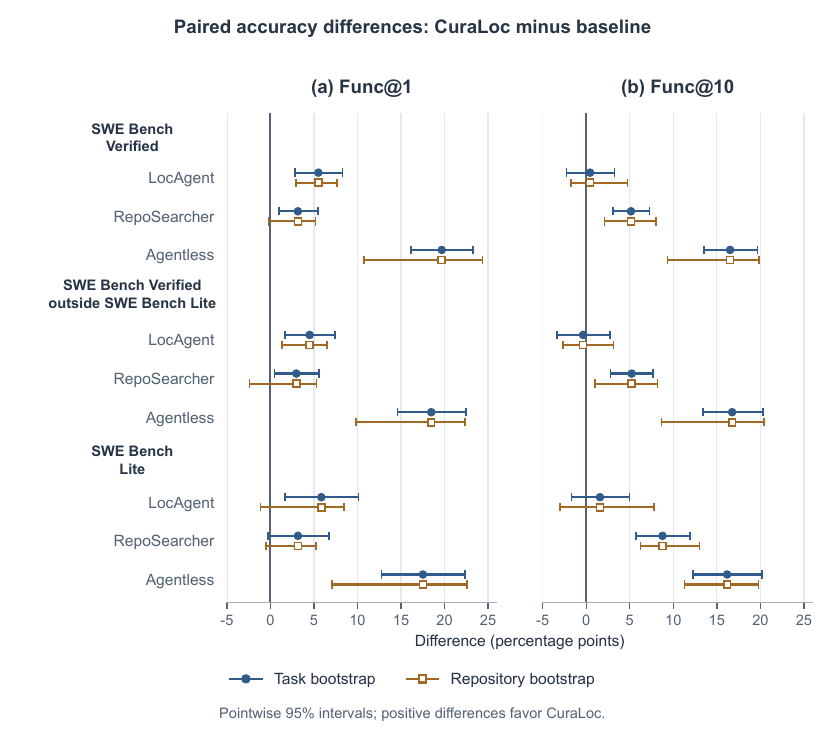}
\caption{CuraLoc-minus-baseline differences in
(a) Func@1 and (b) Func@10 across all nine comparisons.
Circles and squares show pointwise 95\% task- and
repository-bootstrap intervals, respectively.
Positive differences favor CuraLoc. Intervals are
unadjusted for multiple comparisons.
RepoSearcher returns at most five symbols, so its
Func@10 equals Func@5. The non-overlapping cohort
is a subset of SWE Bench Verified.}
\label{fig:paired-effects}
\end{figure}

The repository analysis preserves within-repository grouping but is
limited by the small number of repositories. It assesses sensitivity
to the observed repository mix, not performance on unseen codebases.
Figure~\ref{fig:paired-effects} compares task- and repository-bootstrap
intervals.

\paragraph{Joint accuracy--token comparisons.}
The primary Pareto analysis considers two objectives: higher Func@1
and lower recorded tokens per task. For each baseline, we calculate
the accuracy advantage as CuraLoc minus baseline and token savings
as baseline minus CuraLoc. Both quantities are computed using the same
bootstrap draws. Positive values therefore favor CuraLoc on both axes.

For an individual comparison, we use the 2.5th percentile as a
one-sided lower bound for each objective. The two bounds jointly
target 95\% coverage through a Bonferroni adjustment. Strict dominance
is supported when both lower bounds exceed zero.

We adjust across nine comparisons: CuraLoc versus each
of three baselines on SWE Bench Lite, SWE Bench Verified,
and the SWE Bench Verified subset absent from SWE Bench
Lite. This family contains 18 objective bounds.
Each uses lower-tail probability $0.05/18$, approximately
the 0.278th percentile, targeting 95\% simultaneous
coverage under the bootstrap assumptions.
Unavailable Agentless token bounds retain their slots
in the multiplicity adjustment but yield no token
or dominance conclusion.
The same adjustment applies to whole-repository
resamples as a sensitivity analysis.

Token ratios are computed as the baseline mean divided by the
CuraLoc mean within each resample, not as averages of per-task ratios.
Func@10--token comparisons are secondary and descriptive, outside
the primary dominance family.

\paragraph{Reproducibility and interpretation.}
We use NumPy's PCG64 generator with primary seed 20260923,
separate deterministic streams for cohorts and resampling schemes,
and linear interpolation for quantiles. A second seed, 20260924,
checks Monte Carlo stability; it does not replace the primary
analysis. CuraLoc's dominance over LocAgent on full SWE Bench
Verified remains supported with both seeds.
Two classifications change: the family-adjusted
task-bootstrap Func@1 bound for CuraLoc versus
RepoSearcher on SWE Bench Verified, and the pointwise
repository-bootstrap File@10 bound for RepoSearcher
versus Agentless on SWE Bench Lite.
Both lower bounds reach zero with the second seed,
so these classifications are not seed-stable.

The analysis is post hoc. Multiplicity adjustment does not remove
configuration-selection bias from SWE Bench Lite, and an interval containing
zero does not establish equivalence. Statistical dominance refers
only to the specified accuracy and recorded-token objectives.
Source manifests, per-task outcomes, and bootstrap distributions
are retained for reproducibility.
\section{Additional Results Details}
\label{app:results-details}

\subsection{SWE Bench Verified Cohorts}
\label{app:verified-cohorts}

We partition the 454 scoreable SWE Bench Verified instances into 370 absent
from SWE Bench Lite and 84 shared with SWE Bench Lite. Table~\ref{tab:verified-cohorts}
reports results obtained by restricting each of the three
SWE Bench Verified runs to these fixed subsets. The partition is by task,
not repository: the non-overlapping cohort does not constitute
an unseen-repository evaluation. The overlapping cohort is
included for completeness, not as independent generalization
evidence.

\begin{table*}[t]
\centering
\caption{GPT-5.2 localization on fixed SWE Bench Verified subsets.
Entries are means across three runs, with $\pm$ sample standard
deviations in subscripts. Accuracy values are percentages;
MRR is first-hit reciprocal rank on a 0--1 scale.
These subsets reuse the full SWE Bench Verified executions.}
\small
\setlength{\tabcolsep}{3pt}
\begin{tabular}{lrrrrrrr}
\toprule
Method & Func@1 & Func@3 & Func@5 & Func@10
       & MRR & File@1 & File@10 \\
\midrule
\multicolumn{8}{l}{\textbf{SWE Bench Verified outside SWE Bench Lite ($n=370$)}} \\
CuraLoc
    & $55.5_{\pm0.7}$
    & $69.4_{\pm0.3}$
    & $73.9_{\pm0.6}$
    & $76.8_{\pm1.1}$
    & $0.837_{\pm0.006}$
    & $72.9_{\pm0.6}$
    & $86.8_{\pm0.7}$ \\
LocAgent
    & $51.0_{\pm0.4}$
    & $67.7_{\pm0.7}$
    & $73.1_{\pm0.4}$
    & $77.1_{\pm0.3}$
    & $0.797_{\pm0.003}$
    & $72.3_{\pm0.8}$
    & $94.1_{\pm0.8}$ \\
    RepoSearcher
    & $52.5_{\pm0.4}$ & $66.7_{\pm0.6}$
    & $71.5_{\pm1.0}$ & $71.5_{\pm1.0}$
    & $0.793_{\pm0.001}$ & $73.8_{\pm0.7}$ & $87.5_{\pm0.9}$ \\
Agentless
    & $37.0_{\pm1.0}$ & $53.8_{\pm1.4}$
    & $58.4_{\pm2.0}$ & $60.0_{\pm1.6}$
    & $0.605_{\pm0.023}$ & $67.7_{\pm1.2}$ & $79.9_{\pm0.8}$ \\
\midrule

\multicolumn{8}{l}{\textbf{SWE Bench Verified overlapping SWE Bench Lite ($n=84$)}} \\
CuraLoc
    & $76.2_{\pm4.3}$
    & $88.5_{\pm1.4}$
    & $91.7_{\pm1.2}$
    & $92.1_{\pm1.4}$
    & $0.850_{\pm0.030}$
    & $88.9_{\pm1.4}$
    & $95.2_{\pm1.2}$ \\
LocAgent
    & $66.3_{\pm0.7}$
    & $78.2_{\pm1.4}$
    & $82.5_{\pm0.7}$
    & $88.1_{\pm1.2}$
    & $0.766_{\pm0.004}$
    & $79.8_{\pm1.2}$
    & $96.0_{\pm0.7}$ \\
    RepoSearcher
    & $72.2_{\pm0.7}$ & $82.1_{\pm1.2}$
    & $87.3_{\pm0.7}$ & $87.3_{\pm0.7}$
    & $0.806_{\pm0.014}$ & $84.9_{\pm1.8}$ & $93.7_{\pm0.7}$ \\
Agentless
    & $51.2_{\pm2.1}$ & $71.0_{\pm4.5}$
    & $75.8_{\pm1.4}$ & $76.6_{\pm1.8}$
    & $0.647_{\pm0.016}$ & $79.8_{\pm1.2}$ & $90.5_{\pm3.6}$ \\
\bottomrule
\end{tabular}
\label{tab:verified-cohorts}
\end{table*}

\subsection{Recorded Token Accounting}
\label{app:timing-details}

We measure recorded input-plus-output tokens in the retained
executions that produce each task's final predictions. For each
run, we sum these tokens over the scoreable cohort and divide
by its task count: 274 for SWE Bench Lite, 454 for SWE Bench Verified, and 370 for
SWE Bench Verified outside SWE Bench Lite. For GPT-5.2, we report the mean and sample standard
deviation across three run-level averages.
For GPT-6 Astra, we report one run per method on
the same 454 scoreable tasks, without between-run
standard deviations.

\paragraph{Included computation.}
CuraLoc totals include all three primary stages and summary-model
responses contributing to the retained symbol-selection stage.
LocAgent totals include both retained trajectories used to produce
its fused prediction. RepoSearcher totals cover its file- and
function-localization calls. Agentless totals cover file selection,
related-code localization, and its four final localization samples.
Multiple samples contributing to a prediction are counted together,
rather than selecting only the most successful sample.
On SWE Bench Verified, the three-run stage averages are 30.5K
tokens for file selection, 43.1K for summary generation,
30.4K for symbol and action selection, and 21.7K for
final ranking. Stage totals reconcile with published
usage for all 2,184 CuraLoc main task executions across
SWE Bench Lite and SWE Bench Verified.

\paragraph{Caches and checkpoint recovery.}
Main CuraLoc runs disable summary-cache reuse across tasks and runs.
A resumed task may reuse a completed stage from the same execution;
the requests that originally produced that checkpoint are counted
once. Reading a checkpoint neither adds another generation cost
nor removes the original one. Provider-side cached-input tokens
remain part of the reported input total. Cached-input and reasoning
token breakdowns are not added separately when already included
in the provider's input/output totals.

\paragraph{Retained versus abandoned execution.}
Successful-response usage is complete for CuraLoc,
LocAgent, and RepoSearcher on both GPT-5.2 benchmarks,
and for Agentless on SWE Bench Lite.
For Agentless on SWE Bench Verified, retained-stage
usage is missing for two, three, and three prompt-limit
tasks in the respective runs. These tasks remain
accuracy misses, but exact token means, ratios, and
Pareto comparisons involving Agentless on this benchmark
are not reported. Requests belonging to abandoned stage
executions or discarded trajectories are outside the retained-execution
metric. Failed requests are tracked separately where available;
unreported usage is not assigned a measured value of zero.
For CuraLoc, checkpoint receipts and request records identify the
contributing stages. Baseline totals follow their retained
trajectory or successful-call exports.

\paragraph{Ratios and experimental scope.}
A token-reduction ratio is the baseline's three-run mean divided
by CuraLoc's three-run mean, not an average of per-task ratios.
Quality and token measurements use the same executions and cohorts.
The main comparisons are distinct from the cache-assisted development
runs used for hyperparameter selection and ablations. Their
stage-specific token measurements are identified separately and
are not substituted for end-to-end main-comparison totals.

\subsection{Selection, Inspection, and Final Recovery}
\label{app:failure-analysis}

We analyze the three retained GPT-5.2 runs of CuraLoc
and LocAgent on the same 454 SWE Bench Verified tasks.
All final outcomes reproduce the published Func@10
results. The runs repeat the same tasks; pooled counts
describe task--run executions rather than independent
benchmark instances.

\paragraph{Coverage definitions.}
For CuraLoc, coverage means that the symbols selected
for S3 contain every normalized reference edit symbol,
including neighborhood expansion and before the final
input-budget limit. A selected symbol need not have
its complete implementation displayed, since source
windows are bounded and the final source-block lists
were not retained.

For LocAgent, coverage requires the complete source
implementations of all reference symbols to appear
in tool outputs from its two retained trajectories.
Source ranges are resolved against the recorded
repository revision. Partial source, names, signatures,
graph listings, tool arguments, and final predictions
do not establish complete implementation exposure.
Coverage is the union across the two trajectories;
it does not imply that one model context contained
all implementations simultaneously.

These definitions reflect the available records and
are not interchangeable measurements of final-model
context. LocAgent inspection coverage is unknown
for 23, 24, and 31 tasks in the respective runs because
of source/revision mismatches, unresolved symbol
identities, or incomplete observation alignment.
These tasks remain in the accuracy denominator.

\paragraph{Failure categories.}
We classify each Func@10 failure as having incomplete
recorded coverage, complete coverage but incomplete
final recovery, or insufficient records to determine
coverage (Table~\ref{tab:inspection-failures}).
These are observed outcome categories, not isolated
causes. In particular, a CuraLoc miss after complete
selection can include source truncation as well as
an unsuccessful ranking decision.

\begin{table}[t]
\centering
\caption{Func@10 failures on SWE Bench Verified.
A: incomplete coverage under the method-specific
definition; B: complete coverage but incomplete final
recovery; C: unknown coverage.
All 454 tasks remain in each run's denominator.}
\label{tab:inspection-failures}
\small
\setlength{\tabcolsep}{8pt}
\begin{tabular}{@{}llrrrr@{}}
\toprule
Method & Run & A & B & C & Total failures \\
\midrule
CuraLoc & 1 & 94 & 4 & 0 & 98 \\
        & 2 & 79 & 9 & 0 & 88 \\
        & 3 & 88 & 4 & 0 & 92 \\
\midrule
LocAgent & 1 & 23 & 63 & 10 & 96 \\
         & 2 & 23 & 59 & 13 & 95 \\
         & 3 & 20 & 60 & 13 & 93 \\
\bottomrule
\end{tabular}
\end{table}

Of CuraLoc's 278 failed task--run executions, 261
have incomplete selected-symbol coverage (93.9\%).
This category comprises 36, 35, and 32 failures with
reference symbols missing from the S2 candidate set,
and 58, 44, and 56 with complete candidate coverage
but incomplete source selection.
Conversely, 1,077 of the 1,094 executions with complete
selection coverage recover the reference edit set
within ten predictions (98.4\%).

LocAgent has 182 failures despite complete source
exposure, representing 64.1\% of its 284 failures.
Another 36 failures have unknown exposure.
These counts distinguish reaching reference code
from recovering its complete edit set in the final
prediction.

\paragraph{Paired outcomes on jointly covered tasks.}
Table~\ref{tab:joint-coverage-recovery} restricts each
run to tasks where CuraLoc selects every reference
symbol and LocAgent observes every complete reference
implementation. Both methods are evaluated on the
same tasks within each row.

\begin{table}[t]
\centering
\caption{Final recovery on tasks with complete coverage
under both methods' respective definitions.
Counts are Func@10 successes.}
\label{tab:joint-coverage-recovery}
\small
\setlength{\tabcolsep}{8pt}
\begin{tabular}{@{}lrrrr@{}}
\toprule
Run & Tasks & CuraLoc & LocAgent & Difference \\
\midrule
1 & 342 & 338 & 316 & +22 \\
2 & 359 & 351 & 327 & +24 \\
3 & 345 & 341 & 319 & +22 \\
\bottomrule
\end{tabular}
\end{table}

The paired comparison removes differences in task
membership within each row, but does not match the
source presentation, available context, or trajectory
structure. It therefore describes final recovery under
the evaluated methods, rather than isolating the effect
of their ranking procedures.

\paragraph{Successes outside recorded coverage.}
CuraLoc succeeds on three, one, and three tasks despite
incomplete explicit selected-symbol coverage.
LocAgent has no such successes among classifiable
records, but 13, 11, and 18 successful tasks have
unknown coverage. All remain successes.
Other prompt content or surrounding source may support
predictions outside an explicit selected-symbol list;
recorded coverage is therefore not an absolute ceiling
on localization accuracy.

\subsection{Ablation Protocol and Diagnostics}
\label{app:ablation-details}

All comparisons use the same 274 scoreable SWE Bench Lite instances and
a fixed full-CuraLoc reference with $F=10$, $T=3$, and $L=60$.
The reference achieves 67.2\% Func@1 and 87.6\% Func@10.
Each intervention is evaluated separately; removals are not
cumulative. These development experiments are distinct from
the three-run cold evaluations in the main comparison.

\begin{table*}[t]
\centering
\caption{Stage diagnostics, component interventions, and source-depth
comparisons against the fixed SWE Bench Lite reference. Accuracy values are
percentages; MRR uses the first correct symbol. File scoping reports
native file accuracy, whereas other File@1 values are derived from
symbol predictions. Source-depth settings retain the reference
selection-stage outputs.}
\small
\setlength{\tabcolsep}{3pt}
\begin{tabular}{@{}lrrrrrr@{}}
\toprule
Configuration & Func@1 & Func@3 & Func@5 & Func@10
              & MRR & File@1 \\
\midrule
Full CuraLoc (reference)
    & 67.2 & 81.8 & 85.0 & 87.6 & 0.823 & 88.0 \\
\midrule
File scoping only
    & -- & -- & -- & -- & -- & 74.1 \\
Selection-stage readout
    & 55.8 & 76.3 & 79.6 & 83.2 & 0.739 & 80.3 \\
Source-order selection
    & 42.3 & 51.1 & 53.3 & 55.1 & 0.518 & 75.9 \\
\midrule
Without summaries
    & 64.6 & 81.0 & 84.3 & 86.1 & 0.807 & 86.1 \\
Starting symbols only
    & 67.2 & 81.0 & 83.2 & 83.9 & 0.822 & 88.3 \\
Relevance-guided formulation
    & 63.1 & 78.5 & 81.4 & 82.8 & 0.775 & 84.3 \\
\midrule
Source window $L=30$
    & 65.0 & 81.4 & 85.0 & 87.6 & 0.812 & 85.8 \\
Source window $L=100$
    & 66.8 & 81.4 & 85.0 & 87.6 & 0.823 & 87.2 \\
\bottomrule
\end{tabular}
\label{tab:ablation-details}
\end{table*}

\paragraph{Stage diagnostics.}
The reference file-scoping output achieves 74.1\% File@1 and
contains all gold files within its top ten on 269 of 274 tasks
(98.2\%). The selection-stage row is an offline readout of the
saved S2 ranking. Comparing it with full CuraLoc measures the
combined addition of neighborhood expansion, source evidence,
and final ranking, rather than reranking identical evidence.

The source-order intervention retains the reference S1 output,
admitted symbol pool, and issue-independent summaries, but
replaces model-guided S2 selection and guidance with deterministic
selection by file rank and source position. It matches the full
system's evidence-symbol count on each task without reusing
its selected identities. Symbol counts are matched, not token
counts. The intervention produces net losses of 68 Func@1
and 89 Func@10 successes. All gold symbols occur in the explicit
evidence set on 148 tasks, compared with 248 for full CuraLoc.
This containment diagnostic is not a strict prediction ceiling,
because rendered source context can expose additional symbols.

\paragraph{Summary removal.}
This intervention retains S1 and recomputes S2 and S3 without
generated, cached, or docstring-derived summary annotations.
Comments and docstrings inside retrieved source remain available.
At Func@1, it gains six tasks and loses thirteen, producing
a net loss of seven. Recorded S2+S3 tokens decrease from
51,985 to 37,927 per task. These stage-specific totals exclude
S1 and summary generation.

\paragraph{Neighborhood removal.}
This intervention retains the reference S2 output and reruns
S3 using only starting-point source code. At Func@1, it gains
one task and loses one; at Func@10, it gains one and loses
eleven. S3 tokens decrease from 22,892 to 6,545 per task.
Combining the new S3 usage with the retained S2 usage gives
35,638 S2+S3 tokens, versus 51,985 for the reference.
This is a reconstructed stage total, not a fresh execution
of both stages.

\paragraph{Full-pipeline relevance guidance.}
The relevance-guided variant recomputes all three stages.
S1 replaces ownership, execution, and enforcement perspectives
with issue-surface, execution-path, and dependency-context
perspectives. Its file annotations describe direct or contextual
relevance rather than consumer/mechanism status.

S2 replaces the responsibility-role vocabulary with
\begin{center}
\begin{tabular}{@{}ll@{}}
\texttt{ISSUE\_INTERFACE}, & \texttt{EXECUTION\_STEP},\\
\texttt{CONTROL\_FLOW\_CONTEXT}, & \texttt{DATA\_FLOW\_CONTEXT},\\
\texttt{STATE\_OR\_CONFIG\_CONTEXT}, & \texttt{VALIDATION\_CONTEXT},\\
\texttt{CALL\_RELATION\_CONTEXT}, and & \texttt{INHERITANCE\_CONTEXT}.
\end{tabular}
\end{center}
Scores describe the usefulness of a symbol's evidence for
localizing the required edit. Explanations, starting points,
investigation actions, and causal hypotheses remain.
S3 receives these relevance-oriented annotations and continues
to rank symbols most likely to require modification.

Both formulations retain the same models, structural operations,
summarization policy, and configured file and source limits.
Selected files, evidence identities, and realized token consumption
may differ. The experiment therefore compares complete guidance
formulations, not ranking over identical context.

The relevance run uses separate stage artifacts and a private
copy of an existing summary cache. Matching entries are reused,
and missing descriptions are generated normally. The seed does
not by itself establish identical summary text to that seen
by every task in the historical reference. Artifact hashes and
completion receipts verify the recorded run lineage.
Relative to the reference, the variant has 11 fewer Func@1
successes and 13 fewer Func@10 successes. Its first-hit MRR
is recomputed from normalized predictions using the same
definition as the other rows.

\section{Hyperparameter Sweep Details}
\label{app:hp-sweep}

Table~\ref{tab:hp-sweep} reports the configurations underlying
Figure~\ref{fig:hp-sweep}. All settings use the same 274 scoreable
SWE Bench Lite instances.

\paragraph{File-policy sweep.}
Stage~A varies candidate-file count $F\in\{5,10,15\}$ and
fully admitted top-file count $T\in\{1,3,5\}$ with $L=60$.
File-scoping outputs are held fixed, while symbol selection
and final ranking are recomputed. Available issue-independent
summaries are reused. The selected $F=10,T=3$ setting averages
three development runs; each alternative has one run.

\paragraph{Source-depth sweep.}
Stage~B fixes $F=10,T=3$ and varies $L\in\{30,60,100\}$.
It retains the reference run's S2 output and ordered evidence-symbol
set, changing only the source-window limit used by S3.
The $L=60$ row reuses that reference's existing result.
Consequently, the selected rows in Stages~A and~B differ because
one averages three runs and the other uses one fixed run.

\begin{table*}[t]
\centering
\caption{Hyperparameter sensitivity on SWE Bench Lite ($n=274$).
Bold identifies the selected configuration, not the largest
value in each column. Subscripts show sample standard deviations
for the replicated setting. Token counts are means over tasks
and cover different stages in the two panels; both exclude
S1 and summary generation.
$^{\dagger}$The $F=5,T=1$ setting retains one deterministic
empty-pool miss and its 1,375 recorded tokens without stage
attribution, contributing approximately five tokens per task.}
\small
\setlength{\tabcolsep}{3pt}
\begin{tabular}{@{}rrrrrrrrrr@{}}
\toprule
$F$ & $T$ & $L$ & Runs & Func@1 & Func@3 & Func@5
    & Func@10 & MRR & Tokens/task \\
\midrule
\multicolumn{10}{l}{\textbf{Stage A: file policy; tokens cover S2+S3}} \\
5 & 1 & 60 & 1
    & 59.1 & 73.4 & 75.9 & 77.4 & 0.730 & 29,636$^{\dagger}$ \\
5 & 3 & 60 & 1
    & 64.2 & 79.9 & 83.2 & 85.4 & 0.797 & 41,461 \\
5 & 5 & 60 & 1
    & 62.8 & 78.5 & 81.8 & 85.8 & 0.784 & 46,635 \\
10 & 1 & 60 & 1
    & 58.8 & 73.7 & 77.0 & 78.5 & 0.731 & 38,131 \\
\textbf{10} & \textbf{3} & \textbf{60} & \textbf{3}
    & $\mathbf{65.5}_{\pm1.8}$
    & $\mathbf{80.4}_{\pm1.3}$
    & $\mathbf{84.2}_{\pm0.8}$
    & $\mathbf{86.7}_{\pm0.8}$
    & $\mathbf{0.807}_{\pm0.014}$
    & \textbf{51,352} \\
10 & 5 & 60 & 1
    & 65.7 & 81.4 & 85.0 & 88.0 & 0.806 & 56,886 \\
15 & 1 & 60 & 1
    & 59.1 & 73.0 & 77.4 & 79.6 & 0.731 & 39,792 \\
15 & 3 & 60 & 1
    & 63.5 & 80.7 & 84.7 & 86.9 & 0.795 & 57,850 \\
15 & 5 & 60 & 1
    & 67.5 & 83.2 & 85.8 & 87.2 & 0.826 & 65,687 \\
\midrule
\multicolumn{10}{l}{\textbf{Stage B: source depth; tokens cover S3 only}} \\
10 & 3 & 30 & 1
    & 65.0 & 81.4 & 85.0 & 87.6 & 0.812 & 21,136 \\
\textbf{10} & \textbf{3} & \textbf{60} & \textbf{1}
    & \textbf{67.2} & \textbf{81.8} & \textbf{85.0}
    & \textbf{87.6} & \textbf{0.823} & \textbf{22,892} \\
10 & 3 & 100 & 1
    & 66.8 & 81.4 & 85.0 & 87.6 & 0.823 & 23,863 \\
\bottomrule
\end{tabular}
\label{tab:hp-sweep}
\end{table*}

\paragraph{Additional trade-offs.}
Relative to the selected Stage~A mean, $F=10,T=5$ improves
Func@10 by 1.2 percentage points while increasing S2+S3 tokens
by 10.8\%. The broader $F=15,T=5$ setting improves Func@1
by 2.1 points and Func@10 by 0.5 points, at 27.9\% more tokens.
Conversely, $F=5,T=3$ saves 19.3\% of stage tokens while
reducing Func@1 by 1.2 points and Func@10 by 1.3 points.
Differences are calculated before rounding.

With S2 fixed, reducing $L$ from 60 to 30 saves 7.7\% of
S3 tokens; increasing it to 100 adds 4.2\%.
Func@5 and Func@10 are unchanged across these source depths.
These are stage-specific development measurements, distinct
from the complete cold-run token totals used for the main
method comparison.

\end{document}